\documentclass[10pt,twocolumn,letterpaper]{article}

\usepackage[pagenumbers]{cvpr} 

\definecolor{cvprblue}{rgb}{0.21,0.49,0.74}
\usepackage[pagebackref,breaklinks,colorlinks,allcolors=cvprblue]{hyperref}

\usepackage{comment}
\usepackage{algorithm}
\usepackage{algpseudocode}

\def\paperID{8798} 
\def\confName{CVPR}
\def\confYear{2025}

\title{LayerDropBack: A Universally Applicable Approach for Accelerating Training of Deep Networks}

\author{
    Evgeny Hershkovitch Neiterman  and  Gil Ben-Artzi \\
    School of Computer Science, Ariel University\\
    Ariel, Israel\\
    {\tt\small neiterman@ariel.ac.il \quad https://gil-ba.com}
}

\begin{document}
\maketitle
\begin{abstract}
Training very deep convolutional networks is challenging, requiring significant computational resources and time. Existing acceleration methods often depend on specific architectures or require network modifications. We introduce LayerDropBack (LDB), a simple yet effective method to accelerate training across a wide range of deep networks. LDB introduces randomness only in the backward pass, maintaining the integrity of the forward pass, guaranteeing that the same network is used during both training and inference. LDB can be seamlessly integrated into the training process of any model without altering its architecture, making it suitable for various network topologies. Our extensive experiments across multiple architectures (ViT, Swin Transformer, EfficientNet, DLA) and datasets (CIFAR-100, ImageNet) show significant training time reductions of 16.93\% to 23.97\%, while preserving or even enhancing model accuracy. Code is available at \url{https://github.com/neiterman21/LDB}.
\end{abstract}

\section{Introduction}
\label{sec1}

Deep neural networks (DNNs) have achieved remarkable success in computer vision tasks, consistently delivering state-of-the-art performance across a wide range of applications~\cite{he2016deep, neiterman2024channeldropback, helor, aharon}. Nevertheless, resource-intensive training and deployment of very deep networks remain significant challenges, often requiring substantial time and computational power~\cite{ofir2022smm}. Reducing training time is therefore crucial, as it significantly enhances the efficiency of the development and deployment process. This is especially important in environments where shared computational resources are limited. 


Existing methods, such as dropout~\cite{srivastava2014dropout} and DropBlock~\cite{ghiasi2018dropblock}, have been introduced to improve the training of deep networks. These methods primarily aim to enhance generalization rather than explicitly accelerate training and require modifications to the original architecture. Other approaches, such as Stochastic Depth~\cite{huang2016deep}, skip layers during both forward and backward passes, proving effective for regularization while also reducing training time. However, their applicability is limited, as they are tailored for ResNet architecture and rely on identity skip connections where the input and output dimensions are identical. This makes them difficult to integrate into a wide range of common models. For instance, in U-Net architectures, dropping layers disrupts the critical skip connections that preserve spatial information. Similarly, in Transformers, randomly dropping attention layers or feed-forward layers interferes with the essential cross-attention mechanisms and positional encoding. This raises a natural question: can the training of deep learning models be accelerated regardless of their architecture? Achieving this would enable a unified acceleration approach that could be seamlessly applied across different existing and even future model types.

\begin{figure}
    \centering
    \includegraphics[height=0.65\linewidth, width=0.8\linewidth]{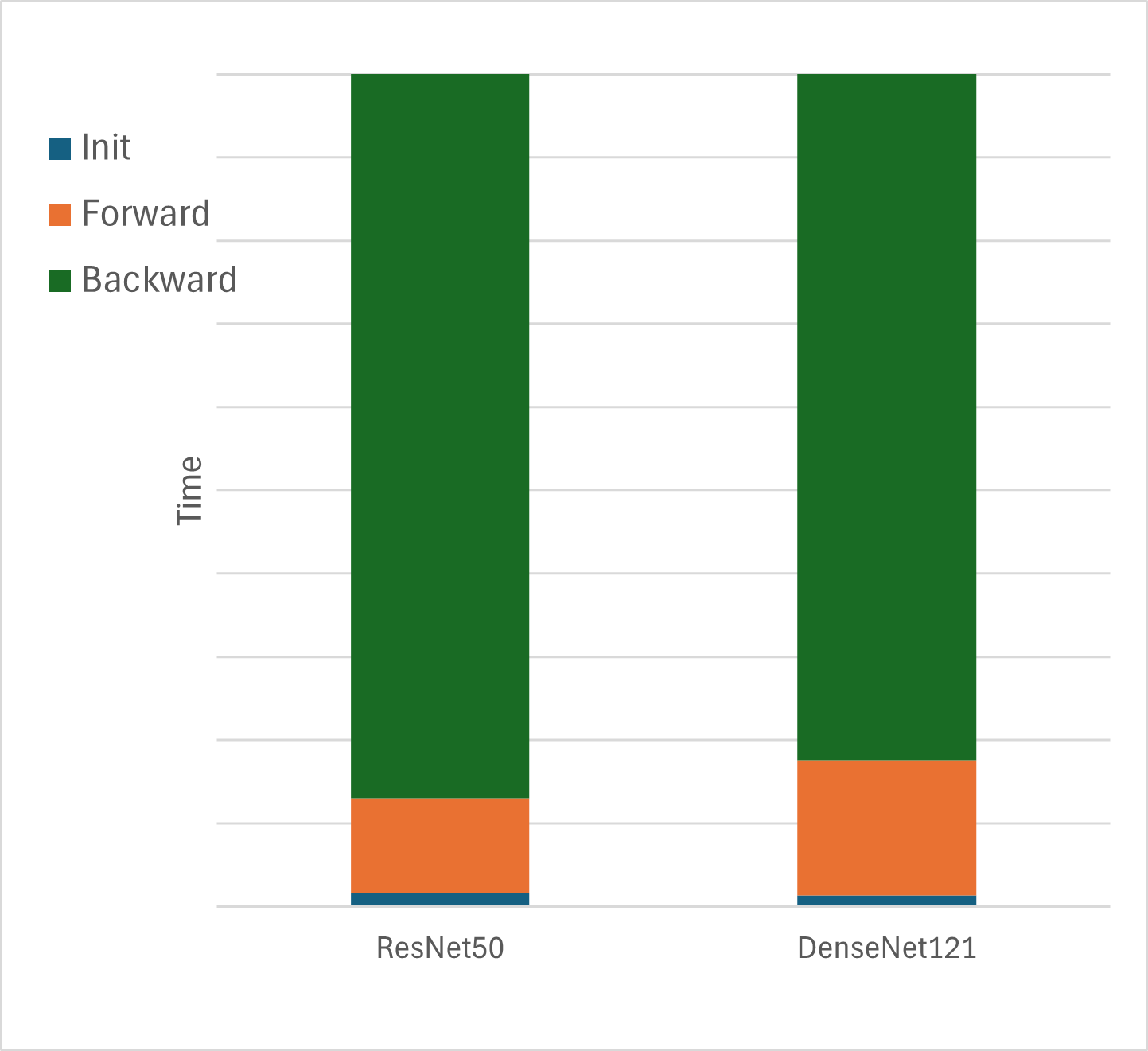}
    \caption{The backpropagation stage dominates training time, consuming 87.2\% for ResNet50 and 82.3\% for DenseNet121.}
    \label{fig:back_cost}
\end{figure}

To this end, we present LayerDropBack (LDB), a universally applicable approach that does not depend on specific architectural designs. LDB accelerates the training of deep learning models by introducing randomness solely into the backward pass, thereby reducing the computational cost of backpropagation — the most expensive phase of training (Figure \ref{fig:back_cost}) — while preserving the integrity of the forward pass. As a result, the network used during training remains identical to the one deployed during inference, without any modifications. Several works aim to reduce the cost of the backpropagation phase~\cite{xiao2019fast}, but they often rely on assumptions regarding layer properties (e.g., the mean of gradients). These methods require additional calculations in every iteration and are only effective when specific criteria are met, limiting their applicability.

Recent research indicates that layer-wise gradients are crucial for efficient optimization, with each layer's gradients being equally important regardless of their specific properties (e.g., ~\cite{Tang_2021_CVPR}). Building on this insight, LayerDropBack incorporates three key components. First, it employs stochastic backpropagation, where a subset of layers is dropped, and gradients are updated based on random subsets of the training data. This can be implemented seamlessly with minimal overhead, enabling faster training. However, to mitigate potential instability from high variance due to repeated stochastic selections, LayerDropBack alternates between epochs that use stochastic backpropagation with a subset of layers and epochs that use standard mini-batch SGD with full backpropagation. This policy can be referred to as semi-stochastic since the training process is neither fully stochastic (as it would be if a subset of layers were always used) nor fully deterministic (as in standard SGD). Instead, it combines both approaches in a structured manner to stabilize training while still reducing overall computational load. Finally, during stochastic backpropagation iterations, LayerDropBack increases both the batch size and learning rate, further stabilizing and accelerating the learning process.

We demonstrate that our approach achieves significant training speedup across various architectures, making it a simple and general-purpose method without compromising accuracy. We evaluate our approach on diverse models such as ViT~\cite{dosovitskiy2020image}, SwinTransformer~\cite{liu2021swin}, EfficientNet~\cite{tan2019efficientnet}, and DLA~\cite{yu2018deep}, across multiple datasets, including CIFAR-100~\cite{krizhevsky2009learning} and ImageNet~\cite{deng2009imagenet}, and across multiple GPUs. Our experimental results reveal significant training time reductions, with mean speedups ranging from 16.93\% to 23.29\%, while maintaining or even slightly improving model accuracy. This results in considerable savings in training time.

\section{Related Work}\label{sec2}

Various related works have been introduced to improve the efficiency of the training. Dropout~\cite{srivastava2014dropout} is a widely adopted regularization technique that randomly drops individual neurons during training with a specified probability, effectively preventing co-adaptation of neurons and improving generalization. By introducing noise to the network, dropout forces the model to learn more robust representations. DropConnect~\cite{wan2013regularization} generalizes dropout by randomly dropping weights instead of activations. DropBlock~\cite{ghiasi2018dropblock} extends dropout by dropping contiguous regions in feature maps instead of individual units. This structured regularization mimics occlusion and enhances generalization in convolutional networks. Expressive dropout \cite{molchanov2019importance} is a method that dynamically adjusts the dropout rate during training based on the importance of each neuron. This allows the network to focus on the most important neurons while still benefiting from the regularization effect of dropout. Expressive dropout has been shown to improve performance on various deep learning tasks, including image classification and language modeling. Although effective in improving generalization, these methods do not explicitly aim to reduce training time.

Deep Networks with Stochastic Depth \cite{huang2016deep} introduces a training procedure that enables shorter training time by skipping layers with a certain probability in the training process. This approach reduces both forward and backward computations but requires skip connections with matching input and output dimensions, and has been primarily validated in the context of ResNet models. FractalNet~\cite{larsson2016fractalnet} introduces a hierarchical, self-similar structure with DropPath regularization technique that randomly drops paths between nodes during training, reducing the computational cost of training by skipping over certain paths in the network. This approach also improves the accuracy but is only applicable to the specific architecture of FractalNet. SkipNet~\cite{wang2018skipnet} employs reinforcement learning to learn a policy for selectively skipping layers in neural networks. Although effective, the need for a learned policy makes it computationally expensive.

Several works were introduced for accelerating the computation by reducing the backpropagation cost. Selective-Backprop \cite{jiang2019accelerating} proposes a technique that accelerates training by skipping over the backpropagation on examples with low loss, under the assumption that examples with low loss contribute little to the gradient update.  \cite{xiao2019fast} is another accelerating  method, which speeds up training by excluding layers from the backward pass. The excluded layers are sampled based on their mean gradient magnitude. Layers with low mean gradient magnitude are more likely to be sampled assuming they contribute less to the optimization, contrary to our stochastic sampling approach. MeProp~\cite{sun2017meprop} demonstrated acceleration of training shallow models with one or two hidden layers by updating only a sparse subset of the inputs of hidden layers during the backpropagation process. Here, we are mainly concerned with commonly used large-scale models and datasets.

\begin{figure}[t]
  \centering
  \includegraphics[width=0.95\linewidth]{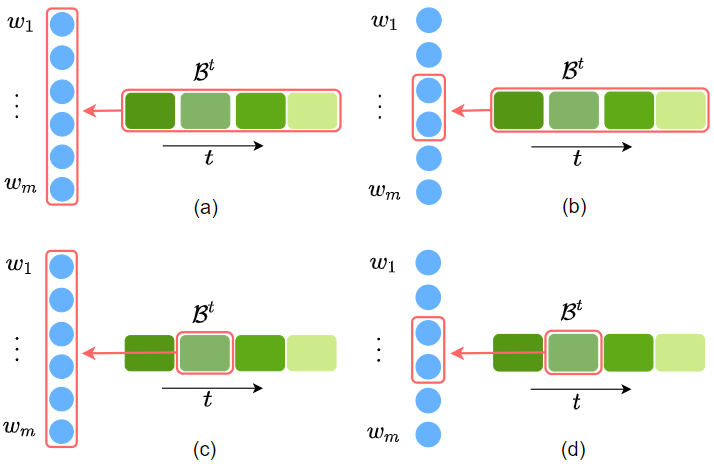}
  \caption{Our approach is semi-stochastic in the parameter space and stochastic in the sample space. $w_1 \ldots w_m$ represent the parameter space and $\mathcal{B}^t$ the mini-batches. A red rectangle that contains only a subset of items represents a stochastic sampling, while containing the whole items represents deterministic sampling. Our approach is alternating between (c) and (d).}
  \hfill
  \label{fig:stochastic_space}
\end{figure}

\section{LayerDropBack (LDB)}

LDB aims to provide a balance between full network training and the regularization effects of partial backpropagation. It alternates between standard Stochastic Gradient Descent (SGD) and a modified SGD with layer-wise dropout during backpropagation with updated batch size and learning rate for stabilization of the dropping process.

\subsection{Epoch-based Alternation Strategy}

LDB employs an epoch-based alternation strategy, switching between standard SGD and the layer dropout method. Let $E$ denote the set of all training epochs, and $e$ be a specific epoch. The training mode for each epoch is determined as follows:

\begin{equation}
\text{Mode}(e) =
\begin{cases}
\text{Standard SGD}, & \text{if } e~\bmod~$s$ \neq 0 \text{~~~or } e = 0 \\
\text{DROP}, & \text{if } e~\bmod~$s$ = 0 \text{ and } e > 0,
\end{cases}
\end{equation}

where $s$ is a predefined parameter that determines the frequency of stochastic backpropagation epochs.

\subsection{Layer Selection Strategy}

During LDB epochs, the method randomly selects a subset of layers within the network for which gradients will be computed and weight updates will be applied. The layer selection strategy is based on a predefined dropout rate, ensuring that each layer has a non-zero probability of being selected.

Formally, let $L$ denote the set of all layers in the network, and $l$ be a layer in $L$. The subset of selected layers, $S$, is given by:

\begin{equation}
S = \{l \in L : u_l < p\},
\end{equation}

where $u_l$ is a random number drawn from a uniform distribution $U(0, 1)$ for each layer $l$, and $p$ is the drop rate hyperparameter.

\subsection{Backward Pass Modification}

In the selected epochs where we intend to skip the update of layers, we adjust the backpropagation algorithm to calculate gradients and update weights exclusively for the layers included in set $S$.
The forward pass remains unchanged, ensuring that the full network is used for inference.

Formally, let $w_l$ denote the weights of layer $l$, and $\Delta w_l$ denote the weight updates computed during backpropagation. The updated weights, $w'_l$, are given by:

\begin{equation}
w'_l = w_l + \Delta w_l \cdot I(l \in S),
\end{equation}

where $I(l \in S)$ is an indicator function that returns 1 if layer $l$ is in the subset $S$, and 0 otherwise.

\subsection{Batch Size and Learning Rate Adjustment}

To compensate for the reduced number of parameter updates during stochastic backpropagation epochs, we increase both the learning rate and the batch size. Let $\eta$ be the base learning rate, $B$ be the base batch size, and $\kappa>1$ a chosen scalar. The adjusted learning rate $\eta'$ and batch size $B'$ for each epoch are given by:

\begin{equation}
(\eta', B') = 
\begin{cases}
    (\frac{1}{p} \eta, \kappa B), & \text{if Mode}(e) = \text{DROP} \\
    (\eta, B), & \text{if Mode}(e) = \text{Standard SGD}. 
\end{cases}
\end{equation}

\subsection{Training Algorithm}

Algorithm \ref{alg:LDB} presents the complete training algorithm for LDB.

\begin{algorithm*}
\caption{LayerDropBack (LDB)}
\label{alg:LDB}
\begin{algorithmic}[1]
\State Initialize network parameters $w$
\For{each epoch $e \in E$}
    \State Determine the training mode: Mode$(e)$
    \State Set {\bf learning rate} $\eta'$ and {\bf batch size} $B'$ based on Mode$(e)$
    \If{Mode$(e) = \text{DROP}$}
        \State Select subset of layers $S$
    \EndIf
    \For{each mini-batch of size $B'$}
        \State Perform forward pass
        \State Compute loss
        \If{Mode$(e) = \text{DROP}$}
            \State Compute gradients only for layers in $S$
            \State Update weights only for layers in $S$ using learning rate $\eta'$
        \Else
            \State Compute gradients for all layers
            \State Update weights for all layers using learning rate $\eta'$ 
        \EndIf
    \EndFor
    
    \State Evaluate model on validation set
\EndFor
\end{algorithmic}
\end{algorithm*}

\subsubsection{Backward Layer Skipping} 

LDB introduces stochasticity only in the backpropagation phase, computing gradients for all layers while selectively skipping parameter updates in some layers to reduce the computational burden of backpropagation. During the forward pass, all parameters are evaluated at all times, resulting in no structural changes to the information flow. This is a key strength, as it makes LDB highly model-robust - it can be applied to a wide variety of architectures regardless of their internal design. It does not require identity skip connections (where input and output is of the same dimension) and residual blocks with the same functional purpose, which are typical of ResNet but do not exist in many other architectures, such as Transformers. As a result, we can apply the same training protocol and parameter selection across various models and tasks, whether fine-tuning or training from scratch, as used in the original implementations.

We observed that it is better to reduce stochasticity for the earlier layers for improved convergence of the model. We therefore employ a simple policy of keeping the four initial layers of the model during the layer selection phase, always training them regularly. 

\subsubsection{Implicit Generalization} 
Despite reducing the number of layers updated during the backpropagation phase, training with LayerDropBack often results in comparable accuracy to training with SGD. This can be explained by the implicit generalization characteristic of our method. Similar to dropout \cite{srivastava2014dropout}, our approach can be considered as a stochastic regularization technique. For a model with $L$ layers, it is equivalent to sampling one of $2^{L}$ possible subsets for which to perform a gradient update in the current iteration. The uncertainty of the descent path helps the optimizer to avoid the problems of bad local minima and saddle points.

\subsection{Inference}

One of the key advantages of LayerDropBack is that it ensures the exact same network is deployed during both training and inference. This is because LayerDropBack does not introduce any modifications to the forward pass or the network architecture during training. Consequently, the inference procedure with LayerDropBack is identical to the standard inference process, without any additional computations or modifications.

\section{Experimental Setup}
\label{sec:experiments}

\subsection{Datasets}
LDB was evaluated on CIFAR-100 (60,000 32x32 color images; 100 classes; 50k train, 10k test), CIFAR-10 (60,000 32x32 color images; 10 classes; 50k train, 10k test). We also use the ILSVRC-2012 ImageNet dataset, referred to as ImageNet1K (1.3M train, 50k validation images; 1000 classes; 224x224 pixels), with its larger superset, ImageNet21K (14.2M+ images; 21,841 classes; varying resolutions, resized to 224x224 pixels)~\cite{deng2009imagenet}.

\subsection{Models} 
LDB was fine-tuned on ViT Large (307M parameters) and Base (86M parameters) \cite{dosovitskiy2020image} and SwinTransformer (88M params) \cite{liu2021swin}, pretrained on ImageNet21K. For evaluating the full training cycle we use MobileNetV2 (3.4M params) \cite{sandler2018mobilenetv2}, ShuffleNetV2 1.0x (2.3M params) \cite{ma2018shufflenet}, EfficientNet-B0 (5.3M params) \cite{tan2019efficientnet}, DLA-34 (15.8M params) \cite{yu2018deep}, ResNet-50 (25.6M params) \cite{he2016deep}, and DenseNet-121 (8M params) \cite{huang2017densely}. These models represent a wide range of sizes and architectures.

\subsection{Training}

\subsubsection{Training Configuration}
SGD with momentum (0.9) was used as the optimizer. For training from scratch, the base learning rate was 0.1 with cosine annealing and batch size was 128. For fine-tuning we use a learning rate of 0.01 and batch size of 512 for ImageNet.
 
\subsubsection{Hardware and Software}
The models were trained on eight Tesla V100 SXM2 32Gb and eight Quadro RTX6000 GPUs, using PyTorch 2.1.2 \cite{NEURIPS2019_9015}, CUDA 12.3, and cuDNN 8.9.2.

\subsubsection{Training Procedure}
\begin{enumerate}
    \item \textbf{Baseline Training}: Each model was first trained using standard Stochastic Gradient Descent (SGD) without any modifications, serving as the baseline for comparison.
    \item \textbf{LDB Training}: The same models were then trained using the LDB method. This involved alternating between standard SGD epochs and epochs with layer dropping as per the LDB methodology. The last layer as well as the first four layers are excluded from the sampling and are always updated.
    \item \textbf{Hyperparameters}: Initial learning rates and batch sizes were set according to best practices for each model and dataset combination. During LDB epochs, the learning rate and batch size were adjusted by a factor of $\kappa = 2$. For all experiments we used $p=0.3$. The same values for $\kappa$ and $p$ were used across all experiments, demonstrating their robustness.
\end{enumerate}

\begin{figure}[tb]
\centering
\includegraphics[width=0.45\textwidth]{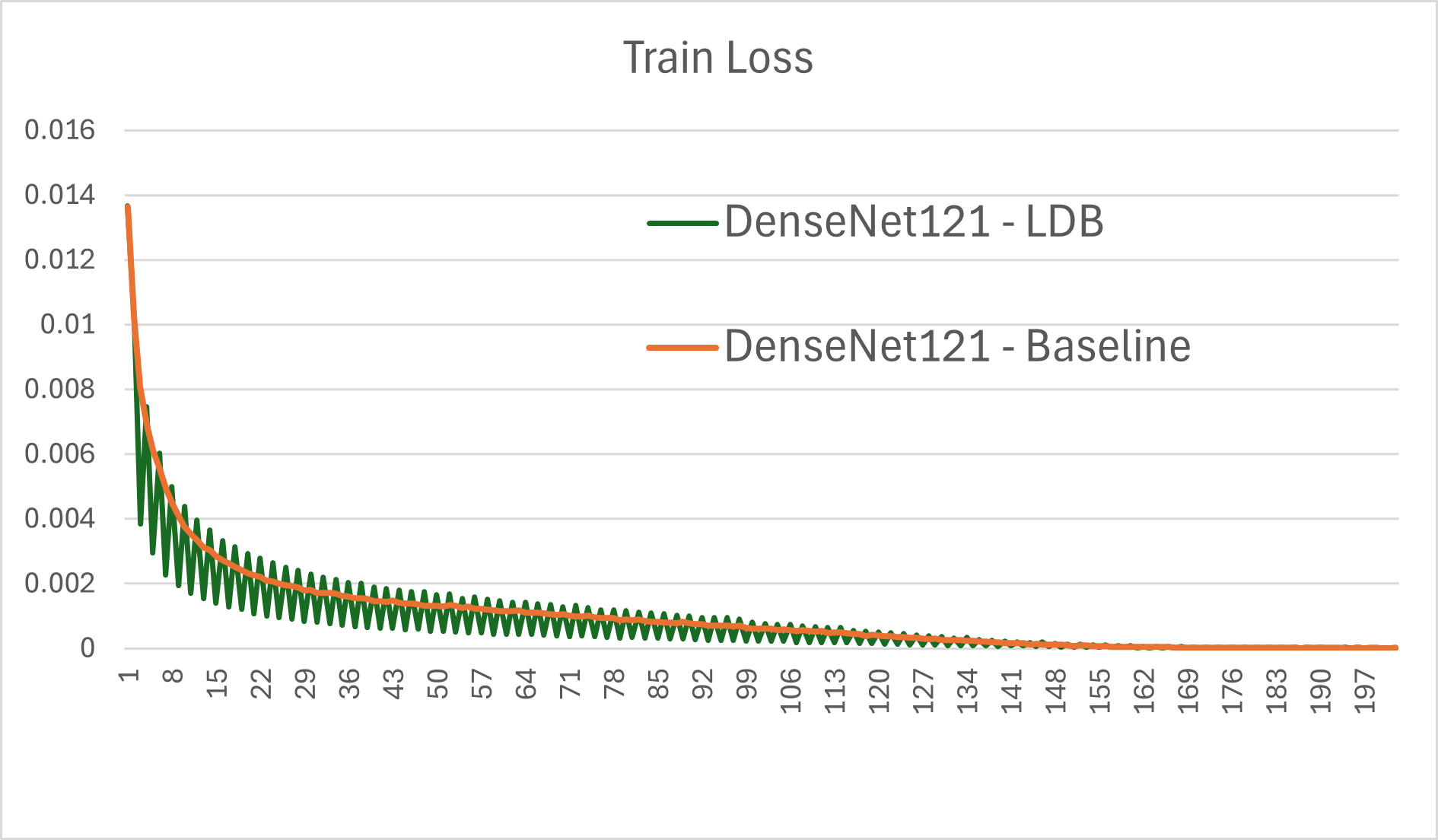} 
\caption{Training loss curves for DenseNet121 on the CIFAR-10 dataset.}
\label{fig:loss_curves}
\end{figure}

\Cref{fig:loss_curves} presents the training loss for both our method, LDB, and the baseline.

\section{Experiments}

We evaluate both the training time speedup and the obtained accuracy for several common models and datasets. We also test the impact of the droprate, alternation frequency, and consistency of performance improvements across different hardware setups.

\subsection{Fine-Tuning}

\Cref{tab:transfer_learning} presents the results of fine-tuning models. The results demonstrate the effectiveness of the LDB method across various models and datasets. For the ViT-Base/16 model, LDB consistently matches or outperforms the baseline in accuracy while providing speedups of 11.8\% to 14.5\%. The larger ViT-Large/16 model shows even more substantial speedups of 21.7\% to 22.5\%, with accuracy improvements on CIFAR-10 and CIFAR-100, but a slight decrease on ImageNet. This suggests that LDB's benefits may scale with model size, offering greater efficiency gains for larger architectures. The SwinTransformer-Base results on ImageNet align with this trend, showing both accuracy improvement and significant speedup. Notably, LDB's performance gains are more pronounced on complex datasets like CIFAR-100 and ImageNet compared to the simpler CIFAR-10, indicating its particular value for challenging tasks. The consistency of LDB across different architectures highlights its versatility. While there is a minor accuracy trade-off for ViT-Large/16 on ImageNet, the substantial speedup likely outweighs this small decrease for many practical applications. 

Overall, in fine-tuning tasks, LDB either maintains or improves accuracy while reducing training time, and is especially beneficial for larger models and more complex datasets.

\begin{table*}[htb]
\centering
\caption{Fine-tuned model performance on various datasets}
\label{tab:transfer_learning}
\begin{tabular}{lllcccc}
\hline
Model & Dataset & Method & Speedup & Accuracy & Accuracy Difference  \\
\hline
ViT-Base/16 & CIFAR-10 & LDB & 11.8\% & \textbf{98.77\%} & 0.00\%  \\
 &  & Baseline &  & \textbf{98.77\%} & \\
 & CIFAR-100 & LDB & 12.0\% & \textbf{92.71\%} & 0.97\%  \\
 &  & Baseline &  & 91.74\% &  \\
 & ImageNet & LDB & 14.5\% & \textbf{81.48\%} & 0.16\%  \\
 &  & Baseline &  & 81.32\% &  \\
ViT-Large/16 & CIFAR-10 & LDB & 21.7\% & \textbf{99.08\%} & 0.04\%  \\
 &  & Baseline &  & 99.04\% & \\
 & CIFAR-100 & LDB & 22.5\% & \textbf{93.48\%} & 0.1\%  \\
 &  & Baseline &  & 93.38\% &  \\
 & ImageNet & LDB & 21.7\% & 82.71\% & -0.23\%  \\
 &  & Baseline &  & \textbf{82.94\%} &  \\
SwinTransformer-Base & ImageNet & LDB & 14.3\% & \textbf{84.46\%} & 2.25\%  \\
 &  & Baseline &  & 82.21\% &  \\
\hline
Mean & & & 16.93\% & & +0.47\%  \\
 \hline
\end{tabular}
\end{table*}

\begin{table*}[htb]
\centering
\caption{Model performance on various datasets trained from scratch}
\label{tab:pref_full_train}
\begin{tabular}{lllclc}
\hline
Model & Dataset & Method & Speedup & Accuracy & Accuracy Difference \\
\hline
DLA & CIFAR-10  & LDB       & 23.6\% & 94.82\%          & -0.24\%   \\
    &           & Baseline  &        & \textbf{95.06\%} &          \\
 & CIFAR-100 & LDB & 24.0\% & \textbf{73.5\%} & 0.36\%  \\
 &  & Baseline &  & 73.14\% &  \\
MobileNetV2 & CIFAR-10 & LDB & 22.5\% & \textbf{94.32\%} & 0.28\%  \\
 &  & Baseline &  & 94.04\% &  \\
 & CIFAR-100 & LDB & 24.5\% & 75.8\% & -0.8\%  \\
 &  & Baseline &  & \textbf{76.6}\% &  \\
EfficientNetB0 & CIFAR-10 & LDB & 21.2\% & 92.58\% & -0.21\%  \\
 &  & Baseline &  & \textbf{92.79\%} &  \\
 & CIFAR-100 & LDB & 21.3\% & 70.72\% & -0.7\%  \\
 &  & Baseline &  & \textbf{71.42\%} &  \\
ShuffleNetV2 & CIFAR-10 & LDB & 24.7\% & \textbf{90.75\%} & 0.64\%  \\
 &  & Baseline &  & 90.11\% &  \\
 & CIFAR-100 & LDB & 23.8\% & 69.36\% & -0.42\% \\
 &  & Baseline &  & \textbf{69.78\%} &  \\
DenseNet121 & CIFAR-10 & LDB & 24.4\% & 95.27\% & -0.16\%  \\
 &  & Baseline &  & \textbf{95.43\%} &  \\
 & CIFAR-100 & LDB & 21.5\% & \textbf{77.92\%} & 0.91\%  \\
 &  & Baseline &  & 77.01\% &  \\
ResNet50 & CIFAR-10 & LDB & 24.3\% & \textbf{93.88\%} & 0.5\%  \\
 &  & Baseline &  & 93.38\% &  \\
 & CIFAR-100 & LDB & 23.8\% & 75.55\% & -0.52\%  \\
 &  & Baseline &  & \textbf{76.07\%} &  \\
Unet & CIFAR-10 & LDB & 23.3\% & \textbf{94.17\%} & 0.05\%  \\
 &  & Baseline &  & 94.12\% & \\
 & CIFAR-100 & LDB & 23.2\% & \textbf{72.56}\% & 0.34\%  \\
 &  & Baseline &  & 72.22\% &  \\
\hline
Mean & & & 23.29\% &  & +0.002\%  \\
\hline
\end{tabular}
\end{table*}

\subsection{Training from Scratch}

Table \ref{tab:pref_full_train} presents the performance results of our proposed LDB approach compared to the baseline for various deep learning models trained from scratch on CIFAR-10 and CIFAR-100 datasets. On average, LDB achieved comparable accuracy to the baseline (+0.002\%) while providing a substantial mean speedup of 23.29\% across all models and datasets, demonstrating that LDB can significantly accelerate training without compromising model performance.

For DLA, LDB showed mixed results with a minor accuracy decrease on CIFAR-10 (-0.24\%) but an improvement on CIFAR-100 (+0.36\%), both with speedups exceeding 23\%. MobileNetV2 saw an accuracy improvement on CIFAR-10 (+0.28\%) and a decrease on CIFAR-100 (-0.8\%), with speedups of 22.5\% and 24.5\%, respectively. EfficientNetB0 experienced accuracy drops on both datasets but maintained speedups above 21\%.

ShuffleNetV2 and ResNet50 showed similar patterns, with improved accuracy on CIFAR-10 but decreases on CIFAR-100, all while achieving speedups between 23\% and 25\%. DenseNet121 saw a small accuracy drop on CIFAR-10 (-0.16\%) but a notable improvement on CIFAR-100 (+0.91\%), with speedups of 24.4\% and 21.5\%, respectively.

For Unet, LDB showed improvements in both CIFAR-10 and CIFAR-100, achieving a speedup of more than 23\% while maintaining or improving accuracy. 

These results demonstrate that LDB consistently provides significant training speedups across various deep learning models and datasets. The minor accuracy trade-offs observed in some instances are generally outweighed by the substantial reductions in training time, and often LDB maintains or even improves accuracy.

\subsection{Impact of Drop Rate on LDB Performance}

\begin{figure}[tb]
\centering
\includegraphics[width=0.5\textwidth]{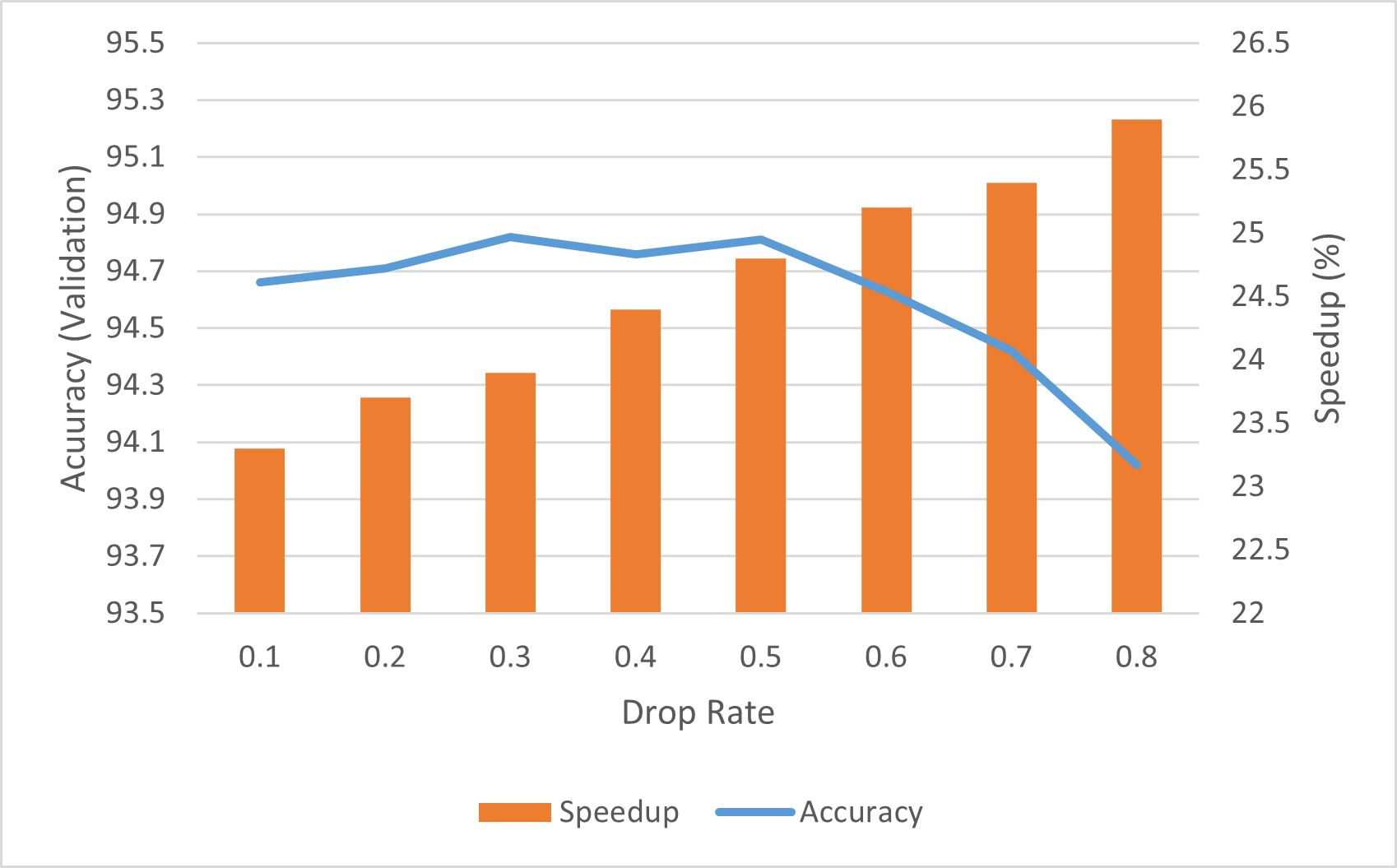}
\caption{Impact of drop rate on top-1 accuracy and speedup for DLA on CIFAR-10.}
\label{fig:drop_rate_impact}
\end{figure}

To investigate the effect of different drop rates on our LDB method, we conducted an ablation study using DLA architecture on the CIFAR-10 dataset (\Cref{fig:drop_rate_impact}).

LDB with a drop rate of 0.1 yields a slight degradation in accuracy to 94.66\% while providing a substantial speedup of 23.3\%. This suggests that a small amount of layer dropping can be beneficial for efficiency without significantly damaging the performance. As the drop rate increases, we observe a general trend of decreasing accuracy, albeit with some fluctuations. The accuracy remains above 94\% for all tested drop rates, demonstrating the robustness of the LDB method. Notably, a drop rate of 0.3 achieves an accuracy of 94.82\%, very close to the baseline, while still maintaining a significant speedup of 23.6\%. The speedup generally increases with higher drop rates, reaching a maximum of 25.7\% at a drop rate of 0.8. However, the relationship between drop rate and speedup is not strictly linear, with some variations observed (e.g., a slight decrease in accuracy from a drop rate of 0.4). The same pattern was observed for various models. 

These results suggest that the choice of drop rate involves a trade-off between accuracy and training speed. Lower drop rates (0.1 to 0.2) tend to preserve accuracy, but offer slightly lower speedups. Higher drop rates (0.6 to 0.8) provide maximum speedup but at the cost of some accuracy. Interestingly, drop rates of 0.3 and 0.5 seem to offer a good balance between accuracy and speedup. 

\begin{table}[tb]
\centering
\caption{Scalability of LayerDropBack for MobileNetV2 on CIFAR100}
\label{tab:scalability}
\begin{tabular}{cc}
\hline
Number of GPUs & Speedup  \\
\hline
1 &  7\%  \\
2 &  18.1\% \\
4 &  21\%  \\
8 &  24.5\%  \\
\hline
\end{tabular}
\end{table}

\begin{table}[tb]
\centering
\begin{tabular}{@{}lcc@{}}
\toprule
Sampling Rate & Accuracy & Speedup  \\
\midrule
1  & 72.82 & 49.3\%  \\
2  & 75.55 & 23.8\%  \\
3  & 75.68 & 14.3\%  \\
4  & 75.82 & 12.3\%  \\
5  & 76.11 & 10.2\%  \\
6  & 75.9 & 7.6\%  \\
7  & 76.73 & 6.5\%  \\
8  & 76.32 & 5.3\%  \\
\bottomrule
\end{tabular}
\caption{Impact of sampling rate on the accuracy and speedup of ResNet50 on CIFAR-100.}
\label{tab:sampling_rate_impact}
\end{table}

\subsection{Scalability}

Table \ref{tab:scalability} shows the speedup achieved by LayerDropBack when scaling from 2 to 8 GPUs for training MobileNetV2 on CIFAR-100. LayerDropBack maintains speedup across different GPU configurations, demonstrating good scalability. Notably, the larger the number of GPUs deployed during training, the more speedup we can achieve. This result was verified on various different types of GPUs (Tesla V100, Quadro RTX6000 GPU, RTX2080Ti).

\subsubsection{Impact of Sampling Rate between Epochs}

\Cref{tab:sampling_rate_impact} shows how model performance varies with different ratios of stochastic to full backpropagation epochs. A sampling rate of 1 indicates that stochastic backpropagation is used for all epochs. Lower sampling rates (less full backpropagation) provide substantial speedups but at the cost of reduced accuracy. Conversely, higher sampling rates enhance accuracy but offer diminishing speedup benefits. The accuracy improvements are notable at a sampling rate around 5 to 7. A sampling rate of 1 yielded the highest speedup at 49.3\%, but this came at the cost of the lowest accuracy, 72.82\%. As the sampling rate continued to rise, the accuracy generally improved, with a sampling rate of 5 achieving an accuracy of 76.11\% and a speedup of 10.2\%. Further increases in the sampling rate beyond 5 resulted in diminishing speedup returns, with rates of 6, 7, and 8 showing speedups of 7.6\%, 6.5\%, and 5.3\%, respectively. The accuracy peaked at a sampling rate of 7 with 76.73\%.

\subsection{Stochastic Backward vs. Forward}

We compare our proposed stochastic backward-only approach to Stochastic Depth (SD)~\cite{huang2016deep}, which introduces stochasticity in both the forward and backward passes. Although SD is effective for regularization and accelerating training, its applicability is mostly restricted to architectures featuring identity skip connections (where input and output dimensions match), primarily ResNet. As a result, our comparison focuses on several ResNet variants (ResNet-20, ResNet-38, and ResNet-56), since SD cannot be easily extended to more diverse architectural designs. We used the recommended implementation by the authors\footnote{\url{https://github.com/felixgwu/img_classification_pk_pytorch}}.

SD achieves comparable accuracy to LDB across all tested ResNet architectures (84.68\% vs 84.64\%). It demonstrates better speedup (28.65\% vs 23.35\%), possibly due to its inherent design which is tightly coupled with the ResNet architecture.

Overall, LDB provides a more architecture-agnostic speedup with promising accuracy benefits in scenarios where broader model compatibility is required, while SD provides a higher speedup for ResNet-only architectures.

\begin{table}[htb]
\centering
\caption{Stochastic Depth and LDB comparison on Resnet18/34/56 on CIFAR datasets}
\label{tab:sd_vs_ldb}
\begin{tabular}{lllcl}
\hline
Model & Dataset & Method & Accuracy &  Speedup \\
\hline
ResNet18 & CIFAR-10 & LDB       & \textbf{94.19}\% & 25\% \\
         &          & SD        & 94.01\% & 28.8\%  \\         
        & CIFAR-100 & LDB       & 74.37\%           & 25\% \\
        &           & SD        & \textbf{74.49}\%           & 28\% \\
ResNet34 & CIFAR-10 & LDB       & 94.6\% & 25\% \\
         &          & SD        & \textbf{94.83}\% & 33\%  \\         
        & CIFAR-100 & LDB       & \textbf{74.87}\%           & 17\% \\
        &           & SD        & 74.86\%           & 33\% \\
ResNet56 & CIFAR-10 & LDB       & \textbf{93.91\%}  & 24.3\% \\
         &          & SD        & 93.82\% &   24.5\%  \\
        & CIFAR-100 & LDB       & 75.95\%           & 23.8\% \\
        &           & SD        & \textbf{76.08}\%           & 24.6\% \\
\end{tabular}
\end{table}

\section{Conclusion}

We presented LayerDropBack (LDB), a simple yet effective method for accelerating deep neural network training. Experiments across various architectures and datasets demonstrated consistent speedups of 16.93\% for fine-tuning and 23.29\% for training from scratch, while maintaining competitive model accuracy (+0.47\% and +0.002\% respectively). The key advantages of LDB are its simplicity, effectiveness, and seamless integration with existing architectures and training pipelines, providing a valuable method for reducing training time and computational costs through layer-wise stochastic backpropagation. While our experiments demonstrated that LDB is effective in computer vision tasks, its principles suggest broader applicability across various domains in deep learning, such as natural language processing (NLP) and speech recognition.

\section{Acknowledgments}
We acknowledge the Ariel HPC Center at Ariel University for providing computing resources that contributed to this research.

{
    \small
    \bibliographystyle{ieeenat_fullname}
    \bibliography{main}

\begin{thebibliography}{26}
\providecommand{\natexlab}[1]{#1}
\providecommand{\url}[1]{\texttt{#1}}
\expandafter\ifx\csname urlstyle\endcsname\relax
  \providecommand{\doi}[1]{doi: #1}\else
  \providecommand{\doi}{doi: \begingroup \urlstyle{rm}\Url}\fi

\bibitem[Aharon and Ben-Artzi(2023)]{aharon}
Shai Aharon and Gil Ben-Artzi.
\newblock Hypernetwork-based adaptive image restoration.
\newblock In \emph{ICASSP 2023 - 2023 IEEE International Conference on Acoustics, Speech and Signal Processing (ICASSP)}, pages 1--5, 2023.

\bibitem[Deng et~al.(2009)Deng, Dong, Socher, Li, Li, and Fei-Fei]{deng2009imagenet}
Jia Deng, Wei Dong, Richard Socher, Li-Jia Li, Kai Li, and Li Fei-Fei.
\newblock Imagenet: A large-scale hierarchical image database.
\newblock In \emph{2009 IEEE conference on computer vision and pattern recognition}, pages 248--255. Ieee, 2009.

\bibitem[Dosovitskiy et~al.(2020)Dosovitskiy, Beyer, Kolesnikov, Weissenborn, Zhai, Unterthiner, Dehghani, Minderer, Heigold, Gelly, et~al.]{dosovitskiy2020image}
Alexey Dosovitskiy, Lucas Beyer, Alexander Kolesnikov, Dirk Weissenborn, Xiaohua Zhai, Thomas Unterthiner, Mostafa Dehghani, Matthias Minderer, Georg Heigold, Sylvain Gelly, et~al.
\newblock An image is worth 16x16 words: Transformers for image recognition at scale.
\newblock \emph{arXiv preprint arXiv:2010.11929}, 2020.

\bibitem[Ghiasi et~al.(2018)Ghiasi, Lin, and Le]{ghiasi2018dropblock}
Golnaz Ghiasi, Tsung-Yi Lin, and Quoc~V Le.
\newblock Dropblock: A regularization method for convolutional networks.
\newblock \emph{Advances in neural information processing systems}, 31, 2018.

\bibitem[He et~al.(2016)He, Zhang, Ren, and Sun]{he2016deep}
Kaiming He, Xiangyu Zhang, Shaoqing Ren, and Jian Sun.
\newblock Deep residual learning for image recognition.
\newblock In \emph{Proceedings of the IEEE conference on computer vision and pattern recognition}, pages 770--778, 2016.

\bibitem[Hel-Or and Ben-Artzi(2021)]{helor}
Yacov Hel-Or and Gil Ben-Artzi.
\newblock The role of redundant bases and shrinkage functions in image denoising.
\newblock \emph{IEEE Transactions on Image Processing}, 30:\penalty0 3778--3792, 2021.

\bibitem[Huang et~al.(2016)Huang, Sun, Liu, Sedra, and Weinberger]{huang2016deep}
Gao Huang, Yu Sun, Zhuang Liu, Daniel Sedra, and Kilian~Q Weinberger.
\newblock Deep networks with stochastic depth.
\newblock In \emph{Computer Vision--ECCV 2016: 14th European Conference, Amsterdam, The Netherlands, October 11--14, 2016, Proceedings, Part IV 14}, pages 646--661. Springer, 2016.

\bibitem[Huang et~al.(2017)Huang, Liu, Van Der~Maaten, and Weinberger]{huang2017densely}
Gao Huang, Zhuang Liu, Laurens Van Der~Maaten, and Kilian~Q Weinberger.
\newblock Densely connected convolutional networks.
\newblock In \emph{Proceedings of the IEEE conference on computer vision and pattern recognition}, pages 4700--4708, 2017.

\bibitem[Jiang et~al.(2019)Jiang, Wong, Zhou, Andersen, Dean, Ganger, Joshi, Kaminksy, Kozuch, Lipton, et~al.]{jiang2019accelerating}
Angela~H Jiang, Daniel L-K Wong, Giulio Zhou, David~G Andersen, Jeffrey Dean, Gregory~R Ganger, Gauri Joshi, Michael Kaminksy, Michael Kozuch, Zachary~C Lipton, et~al.
\newblock Accelerating deep learning by focusing on the biggest losers.
\newblock \emph{arXiv preprint arXiv:1910.00762}, 2019.

\bibitem[Krizhevsky et~al.(2009)Krizhevsky, Hinton, et~al.]{krizhevsky2009learning}
Alex Krizhevsky, Geoffrey Hinton, et~al.
\newblock Learning multiple layers of features from tiny images.
\newblock 2009.

\bibitem[Larsson et~al.(2016)Larsson, Maire, and Shakhnarovich]{larsson2016fractalnet}
Gustav Larsson, Michael Maire, and Gregory Shakhnarovich.
\newblock Fractalnet: Ultra-deep neural networks without residuals.
\newblock \emph{arXiv preprint arXiv:1605.07648}, 2016.

\bibitem[Liu et~al.(2021)Liu, Lin, Cao, Hu, Wei, Zhang, Lin, and Guo]{liu2021swin}
Ze Liu, Yutong Lin, Yue Cao, Han Hu, Yixuan Wei, Zheng Zhang, Stephen Lin, and Baining Guo.
\newblock Swin transformer: Hierarchical vision transformer using shifted windows.
\newblock In \emph{Proceedings of the IEEE/CVF international conference on computer vision}, pages 10012--10022, 2021.

\bibitem[Ma et~al.(2018)Ma, Zhang, Zheng, and Sun]{ma2018shufflenet}
Ningning Ma, Xiangyu Zhang, Hai-Tao Zheng, and Jian Sun.
\newblock Shufflenet v2: Practical guidelines for efficient cnn architecture design.
\newblock In \emph{Proceedings of the European conference on computer vision (ECCV)}, pages 116--131, 2018.

\bibitem[Molchanov et~al.(2019)Molchanov, Mallya, Tyree, Frosio, and Kautz]{molchanov2019importance}
Pavlo Molchanov, Arun Mallya, Stephen Tyree, Iuri Frosio, and Jan Kautz.
\newblock Importance estimation for neural network pruning.
\newblock In \emph{Proceedings of the IEEE/CVF conference on computer vision and pattern recognition}, pages 11264--11272, 2019.

\bibitem[Neiterman and Ben-Artzi(2024)]{neiterman2024channeldropback}
Evgeny~Hershkovitch Neiterman and Gil Ben-Artzi.
\newblock Channeldropback: Forward-consistent stochastic regularization for deep networks.
\newblock In \emph{International Conference on Pattern Recognition}, pages 390--400. Springer, 2024.

\bibitem[Ofir and Ben-Artzi(2022)]{ofir2022smm}
Amir Ofir and Gil Ben-Artzi.
\newblock Smm-conv: Scalar matrix multiplication with zero packing for accelerated convolution.
\newblock In \emph{Proceedings of the IEEE/CVF Conference on Computer Vision and Pattern Recognition}, pages 3067--3075, 2022.

\bibitem[Paszke et~al.(2019)Paszke, Gross, Massa, Lerer, Bradbury, Chanan, Killeen, Lin, Gimelshein, Antiga, Desmaison, Kopf, Yang, DeVito, Raison, Tejani, Chilamkurthy, Steiner, Fang, Bai, and Chintala]{NEURIPS2019_9015}
Adam Paszke, Sam Gross, Francisco Massa, Adam Lerer, James Bradbury, Gregory Chanan, Trevor Killeen, Zeming Lin, Natalia Gimelshein, Luca Antiga, Alban Desmaison, Andreas Kopf, Edward Yang, Zachary DeVito, Martin Raison, Alykhan Tejani, Sasank Chilamkurthy, Benoit Steiner, Lu Fang, Junjie Bai, and Soumith Chintala.
\newblock Pytorch: An imperative style, high-performance deep learning library.
\newblock In \emph{Advances in Neural Information Processing Systems 32}, pages 8024--8035. Curran Associates, Inc., 2019.

\bibitem[Sandler et~al.(2018)Sandler, Howard, Zhu, Zhmoginov, and Chen]{sandler2018mobilenetv2}
Mark Sandler, Andrew Howard, Menglong Zhu, Andrey Zhmoginov, and Liang-Chieh Chen.
\newblock Mobilenetv2: Inverted residuals and linear bottlenecks.
\newblock In \emph{Proceedings of the IEEE conference on computer vision and pattern recognition}, pages 4510--4520, 2018.

\bibitem[Srivastava et~al.(2014)Srivastava, Hinton, Krizhevsky, Sutskever, and Salakhutdinov]{srivastava2014dropout}
Nitish Srivastava, Geoffrey Hinton, Alex Krizhevsky, Ilya Sutskever, and Ruslan Salakhutdinov.
\newblock Dropout: a simple way to prevent neural networks from overfitting.
\newblock \emph{The journal of machine learning research}, 15\penalty0 (1):\penalty0 1929--1958, 2014.

\bibitem[Sun et~al.(2017)Sun, Ren, Ma, and Wang]{sun2017meprop}
Xu Sun, Xuancheng Ren, Shuming Ma, and Houfeng Wang.
\newblock meprop: Sparsified back propagation for accelerated deep learning with reduced overfitting.
\newblock In \emph{International Conference on Machine Learning}, pages 3299--3308. PMLR, 2017.

\bibitem[Tan and Le(2019)]{tan2019efficientnet}
Mingxing Tan and Quoc Le.
\newblock Efficientnet: Rethinking model scaling for convolutional neural networks.
\newblock In \emph{International conference on machine learning}, pages 6105--6114. PMLR, 2019.

\bibitem[Tang et~al.(2021)Tang, Chen, Zhu, Yu, and Ouyang]{Tang_2021_CVPR}
Shixiang Tang, Dapeng Chen, Jinguo Zhu, Shijie Yu, and Wanli Ouyang.
\newblock Layerwise optimization by gradient decomposition for continual learning.
\newblock In \emph{Proceedings of the IEEE/CVF Conference on Computer Vision and Pattern Recognition (CVPR)}, pages 9634--9643, 2021.

\bibitem[Wan et~al.(2013)Wan, Zeiler, Zhang, Le~Cun, and Fergus]{wan2013regularization}
Li Wan, Matthew Zeiler, Sixin Zhang, Yann Le~Cun, and Rob Fergus.
\newblock Regularization of neural networks using dropconnect.
\newblock In \emph{International conference on machine learning}, pages 1058--1066. PMLR, 2013.

\bibitem[Wang et~al.(2018)Wang, Yu, Dou, Darrell, and Gonzalez]{wang2018skipnet}
Xin Wang, Fisher Yu, Zi-Yi Dou, Trevor Darrell, and Joseph~E Gonzalez.
\newblock Skipnet: Learning dynamic routing in convolutional networks.
\newblock In \emph{Proceedings of the European conference on computer vision (ECCV)}, pages 409--424, 2018.

\bibitem[Xiao et~al.(2019)Xiao, Mudiyanselage, Ji, Hu, and Pan]{xiao2019fast}
Xueli Xiao, Thosini~Bamunu Mudiyanselage, Chunyan Ji, Jie Hu, and Yi Pan.
\newblock Fast deep learning training through intelligently freezing layers.
\newblock In \emph{2019 International Conference on Internet of Things (iThings) and IEEE Green Computing and Communications (GreenCom) and IEEE Cyber, Physical and Social Computing (CPSCom) and IEEE Smart Data (SmartData)}, pages 1225--1232. IEEE, 2019.

\bibitem[Yu et~al.(2018)Yu, Wang, Shelhamer, and Darrell]{yu2018deep}
Fisher Yu, Dequan Wang, Evan Shelhamer, and Trevor Darrell.
\newblock Deep layer aggregation.
\newblock In \emph{Proceedings of the IEEE conference on computer vision and pattern recognition}, pages 2403--2412, 2018.

\end{thebibliography}
}

\end{document}